%% file: paper.tex
\documentclass[10pt,twocolumn,letterpaper]{article}
\usepackage{btas}
\usepackage{times}
\usepackage{epsfig}
\usepackage{graphicx}
\usepackage{amsmath}
\usepackage{amssymb}
\usepackage{enumitem}
\usepackage{microtype}
\usepackage{textcomp}  

\usepackage[listofformat=parens, justification=centering, subrefformat=subparens]{subfig}

\btasfinalcopy 

\def\btasPaperID{37} 

\ifbtasfinal\fi

\usepackage{fancyhdr}

\usepackage[pagebackref=true,breaklinks=true,colorlinks,bookmarks=false]{hyperref}

\usepackage{booktabs}

\renewcommand\cap[3]{\caption[#2]{\label{#1}\textsc{#2}. \small#3}}
\usepackage{graphicx}
\graphicspath{{./}{./misc/}{./images/}}
\DeclareGraphicsExtensions{.pdf,.png}

\newcommand\ic[2][1]{\includegraphics[width=#1\textwidth]{#2}}
\renewcommand\etal[1]{\textit{et al.}~\cite{#1}}
\renewcommand\sec[1]{Sec.~\ref{sec:#1}}
\newcommand\fig[1]{Fig.~\ref{fig:#1}}
\newcommand\sfig[1]{Fig.~\subref{fig:#1}}
\newcommand\tab[1]{Tab.\;\ref{tab:#1}}

\DeclareMathOperator\atan{\mathrm{arc\,tan}}

\title{\LARGE \bf
  AFFACT: Alignment-Free\\Facial Attribute Classification Technique
}

\author{Manuel G\"unther,\thanks{Authors contributed equally.}\ \, Andras Rozsa,\footnotemark[1]\ \, and Terrance E. Boult \\
Vision and Security Technology (VAST) Lab \\
University of Colorado, Colorado Springs, USA\\
\small\texttt{\{mgunther,arozsa,tboult\}@vast.uccs.edu}
}

\begin{document}
\maketitle

{
  \chead{\footnotesize This is a pre-print of the original paper accepted for oral presentation at the International Joint Conference on Biometrics (IJCB) 2017.}
  \lhead{}
  \thispagestyle{fancy}
  \pagenumbering{gobble}
}

\begin{abstract}

Facial attributes are soft-biometrics that allow limiting the search space, e.g., by rejecting identities with non-matching facial characteristics such as nose sizes or eyebrow shapes.
In this paper, we investigate how the latest versions of deep convolutional neural networks, ResNets, perform on the facial attribute classification task.
We test two loss functions: the sigmoid cross-entropy loss and the Euclidean loss, and find that for classification performance there is little difference between these two.
Using an ensemble of three ResNets, we obtain the new state-of-the-art facial attribute classification error of 8.00\,\% on the aligned images of the CelebA dataset.
More significantly, we introduce the Alignment-Free Facial Attribute Classification Technique (AFFACT), a data augmentation technique that allows a network to classify facial attributes without requiring alignment beyond detected face bounding boxes.
To our best knowledge, we are the first to report similar accuracy when using only the detected bounding boxes -- rather than requiring alignment based on automatically detected facial landmarks -- and who can improve classification accuracy with rotating and scaling test images.
We show that this approach outperforms the CelebA baseline on unaligned images with a relative improvement of 36.8\,\%.
\end{abstract}


\input{introduction}
\input{relatedwork}
\input{approach}
\input{experiments}
\input{conclusion}

\section*{Acknowledgment}
This research is based upon work funded in part by NSF IIS-1320956 and in part by the Office of the Director of National Intelligence (ODNI), Intelligence Advanced Research Projects Activity (IARPA), via IARPA R\&D Contract No. 2014-14071600012. The views and conclusions contained herein are those of the authors and should not be interpreted as necessarily representing the official policies or endorsements, either expressed or implied, of the ODNI, IARPA, or the U.S. Government. The U.S. Government is authorized to reproduce and distribute reprints for Governmental purposes notwithstanding any copyright annotation thereon.

{\small
\bibliographystyle{ieee}
\bibliography{paper}
}

\end{document}

%% file: introduction.tex
\section{Introduction}

\begin{figure}[t!]
  \centering
  \ic[.229]{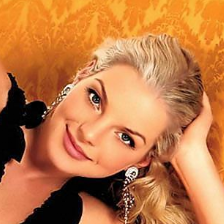}\ \ic[.229]{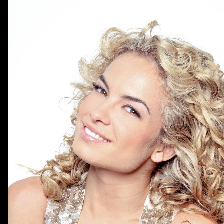}\\[.5ex]
  \ic[.229]{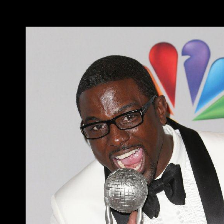}\ \ic[.229]{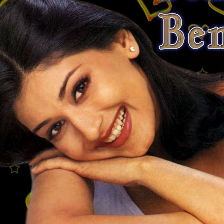}
  \cap{fig:teaser}{Alignment-Free Face Images}{Imagine a deep convolutional neural network that can handle facial images without alignment. Only a dream? Using our data augmentation technique we obtain the AFFACT network, which, for the displayed images, classifies 35 to 40 out of 40 facial attributes correctly!}
\end{figure}

Unlike most other facial features used in recognition tasks, facial attributes have a unique property, namely, a \emph{semantic meaning} that can be interpreted by humans.
Attributes allow descriptive searches (e.g., ``Caucasian female with blond hair'')~\cite{kumar2008facetracer,kumar2011describable,scheirer2012multi} and are interesting for forensic use, i.e., when eyewitnesses describe suspects with such attributes.
After Kumar \etal{kumar2009attribute} showed that facial attributes are an interesting approach for face recognition -- they were state-of-the-art \cite{kumar2011describable} on the Labeled Faces in the Wild (LFW) benchmark \cite{huang2007lfw} prior to the application of Deep Convolutional Neural Networks (DCNNs) -- researchers have shifted their attention towards automatic facial attribute prediction.

Nowadays, DCNNs have become \emph{the quasi-standard} approach for handling almost \emph{any kind of} image recognition or classification task.
This includes generic image classification \cite{lecun1995learning,krizhevsky2012imagenet}, face recognition \cite{taigman2014deepface,parkhi2015deep}, and facial landmark detection \cite{zhang2014landmark,ranjan2016hyperface}.
Also, facial attribute classifiers have been advanced using DCNNs \cite{liu2015deep,rozsa2016adversarial,rudd2016moon,wang2016walk}, mostly leveraging the fact that Liu \etal{liu2015deep} have provided a large-scale facial attribute dataset.
While their algorithm still relies on external training data and additional SVM classifiers \cite{liu2015deep}, it was shown that end-to-end facial attribute prediction systems outperform their baseline without requiring additional training data \cite{rozsa2016adversarial,rudd2016moon}.
Rudd \etal{rudd2016moon} demonstrated that facial attributes extracted with an end-to-end DCNN improve face recognition performance in comparison to the hand-crafted features used by Kumar \etal{kumar2009attribute}.
We assume that this improved face recognition performance is obtained by a better classification of facial attributes.
We hope -- and preliminary experiments show that this hope is justified -- that better attributes improve face recognition accuracy further.

As shown in \cite{dutta14impact,parkhi2015deep}, face recognition works better on aligned faces.
Therefore, for most face processing tasks, the first step is to align the face based on facial landmarks in order to extract features from it \cite{taigman2014deepface,wang2016walk}.
In most of the traditional datasets, face images are high quality and facial landmarks can be detected rather reliably.
However, under more unconstrained face processing -- where no hand-labeled facial landmarks are provided and automatic landmark detection might fail, even for only a few relevant landmarks -- faces cannot be aligned correctly and recognition will fail, even though the face has been detected.
So far, attribute classification approaches in \cite{rozsa2016adversarial,rudd2016moon,wang2016walk} used (hand-annotated) landmark locations while Liu \etal{liu2015deep} did not make use of facial landmarks.
Given that attributes generally cover much larger regions of the image, it is reasonable to ask whether using aligned images is presuming a solution to a harder problem -- facial landmark localization -- in order to solve an easier one, i.e., facial attribute classification.

In this paper, we show that there is no need to align faces because we can use our Alignment-Free Facial Attribute Classification Technique (AFFACT) to reliably predict facial attributes based on faces cropped by using only detected bounding boxes.
For the automatically detected faces shown in \fig{teaser}, our AFFACT network\footnote{Networks of this paper are provided under:\\{\scriptsize\url{http://vast.uccs.edu/public-data/Networks.html}}} is able to predict most of the attributes correctly -- many more than a comparable network.
Data augmentation has been shown to be a valid and useful way to increase the size of the training set without collecting additional training data \cite{howard2014some,chatfield14return}.
Traditionally, data augmentation is performed using simple techniques such as using several crops from an image and their horizontally mirrored versions \cite{howard2014some}.
Other approaches include adding random or non-random noise \cite{rozsa2016hardpositive} and contrast or color modifications \cite{chatfield14return}.
By including such transformations during training, the resulting  DCNNs become more invariant to those transformations.
However, little research has been done on using different kinds of data augmentation such as rotation or scaling \cite{loosli2007training} and no related results have been reported on 2D augmentation for face processing tasks.

In addition,  new network topologies have been developed \cite{he2016deep} that are faster to train and more accurate.
The application of different training regimes \cite{krizhevsky2012imagenet} and loss functions, as well as ensembles of networks can lead to further advancements, but have not been well studied in face applications.
Which of these advances are important for facial attribute prediction, and which of these advances can be exploited by our AFFACT network?
In this paper, we:
\begin{itemize}[topsep=0pt,itemsep=0pt,parsep=0pt]
  \item show that the state of the art in attribute classification can be improved by using the latest trends in DCNNs;
  \item provide an evaluation of two different loss functions;
  \item demonstrate that AFFACT can reliably predict facial attributes from images without alignment; and
  \item propose to use shifts, rotations, and scales of test images to make facial attribute extraction more reliable.
\end{itemize}

%% file: relatedwork.tex
\section{Related Work}

The application of perturbations or distortions to inputs for improving learning models has a long history in machine learning and has become a general training methodology \cite{simard2003best}.
Loosly \etal{loosli2007training} developed an open source tool called InfiMNIST that produces a perturbation-enhanced training set by applying small affine transformations and noise to original samples of the MNIST hand-written digit dataset \cite{lecun1998mnist}.
Krizhevsky \etal{krizhevsky2012imagenet} introduced three types of data augmentation techniques to improve the performance of their network on image classification:
randomly located crops are taken from down-sampled original inputs, training images are horizontally flipped, and intensities of the RGB channels are randomly altered.
Chatfield \etal{chatfield14return} demonstrated that data augmentation methods can also significantly improve the performance of shallow representations.
Baluja \etal{baluja2015virtues} transformed training images such that they were misclassified by a target network but still classified correctly by all \emph{peer} networks.
However, they did not show that training with such examples actually improves performance.
Paulin \etal{paulin2014transformation} used random scaling, rotation, flipping, cropping, and tone change of images in order to increase the training and test sets.
Although our preprocessing technique is similar, we note that in their general object classification task there is no natural alignment for their data, while the possible alignment of faces introduces a new constraint on the training image processing.
Jaderberg \etal{jaderberg2015spatial} trained a spatial transformer network to predict transformation parameters for hand-written digit classification tasks and showed improved accuracy on artificially transformed test images.
While their algorithm required an extension of the classification network and regression of transformation parameters, we neither extend our network, nor do we use the transformation parameters of the images for network training.
Finally, Howard \cite{howard2014some} presented that data transformations during testing can improve DCNN-based image classification.

Only a few papers present results of data augmentation or unaligned faces in face processing tasks, but none of them used the combination of both.
Rowley \etal{rowley1998neural} implemented the first neural network-based face detector, using small perturbations in scale (90-110\,\%), rotation (up to 10\textdegree), and shift (half of a pixel) of the training images to increase the number of training samples.
Although they reported improved face detection results, no further processing of the detected face (e.g., face recognition or attribute extraction) was applied.
Masi \etal{masi2016dowe} rendered images with different facial appearances using 3D models of faces to augment the training set; yet all images were aligned in the center of the training images and upright.
Parkhi \etal{parkhi2015deep} used unaligned images during training, but they only applied simple cropping as data augmentation.
Still, they reported the best face recognition performance on aligned test images.
Maronidis \etal{maronidis2011improving} enhanced their facial expression classification training set using small perturbations in terms of rotation and scaling -- accounting for around 6\,\% of the inter-eye-distance.
For their test images, they still applied alignment based on eye locations.
They reported slightly improved accuracy over training without data augmentation, but they did not try to classify unaligned test images.

Recent approaches to classify facial attributes leverage Deep Convolutional Neural Networks (DCNNs).
Liu \etal{liu2015deep} combined three networks to first localize faces with two Localization Networks (LNets) and then extract features using the Attribute Network (ANet).
Their ANet was first trained with external data for identification and then fine-tuned using facial attributes.
Finally, independent linear Support Vector Machines (SVMs) were used to obtain the final attribute classifications.
Wang \etal{wang2016walk} used an external dataset without manually labeled identities to pre-train a network that was later fine-tuned with attributes, while classification was done using SVMs.
Rozsa \etal{rozsa2016adversarial} trained forty DCNNs -- one for each facial attribute -- on the CelebA dataset and directly used the outputs of the trained networks for facial attribute classification.
This approach does not require secondary classifiers, but still outperforms LNets+ANet by a large margin.
Finally, Rudd \etal{rudd2016moon} introduced the Mixed Objective Optimization Network (MOON) that is able to both classify all facial attributes with a single learning model as well as supporting domain adaptation based on attribute frequency.
This approach is the state of the art on the CelebA dataset.

%% file: approach.tex
\section{Approach}

\subsection{Learning Models and Objective Functions}

Since the current state-of-the-art facial attribute classification on the CelebA dataset was achieved by the MOON architecture \cite{rudd2016moon} built upon the 16-layer VGG model \cite{simonyan15verydeep}, we ask whether applying the latest, more advanced network topologies can lead to further improvements.
The introduction of the residual learning framework by He \etal{he2016deep} allows for training substantially deeper networks than it was previously possible.
The resultant deep Residual Networks (ResNets) are ``easy'' to be optimized -- especially considering their increased depth -- and have achieved excellent performances on various tasks.
For practical considerations, we note that ResNets (with 50, 101, or 152 layers) are faster than VGG models, both during training and testing.

We use two different network topologies, both based on the ResNet-50 model \cite{he2016deep}.
First, we train a new ResNet-50 model using only the CelebA training set while reducing the last fully connected layer to 40 output units -- the number of facial attributes in CelebA.
Later, we investigate whether additional generic training data helps attribute classification.
We start from a ResNet-50 model that was pre-trained on ImageNet for generic image classification.
After adding an additional randomly initialized fully-connected layer with 40 output neurons, we fine-tune that network with the CelebA training set.
We refer to this approach as ResNet-51.

To obtain the state-of-the-art facial attribute recognition results on the CelebA dataset with the Mixed Objective Optimization Network (MOON), Rudd \etal{rudd2016moon} used a Euclidean loss function to mix errors of all facial attributes.
Rozsa \etal{rozsa2016adversarial} also chose a Euclidean loss to train DCNNs -- one for each facial attribute -- following their intuition that ``facial attributes lie along a continuous range, while sigmoids tend to enforce saturation and hinge-loss enforces a large margin.''
Euclidean loss is generally used for regression problems, while facial attribute recognition is a classification task.
Hence, we also apply a more traditional choice for classification tasks, i.e., sigmoid cross-entropy loss, and we evaluate both Euclidean and sigmoid cross-entropy loss on the same dataset using the same network topologies.

\subsection{Data Augmentation for Training}
\label{sec:jittering}

\begin{figure*}
  \centering
  \subfloat[Without Perturbations\label{fig:jittered:no}]{%
    \centering \ic[.17]{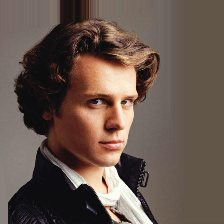}\quad\ic[.17]{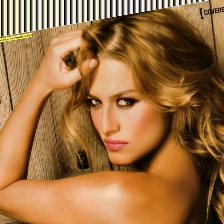}\quad\ic[.17]{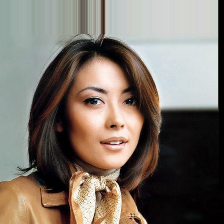}\quad\ic[.17]{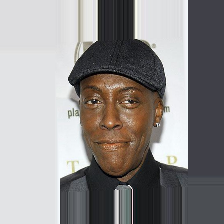}\quad\ic[.17]{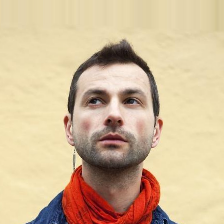}%
  }\\
  \subfloat[With Random Perturbations\label{fig:jittered:yes}]{%
    \centering \ic[.17]{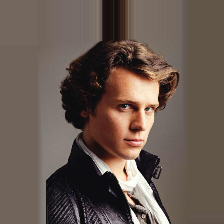}\quad\ic[.17]{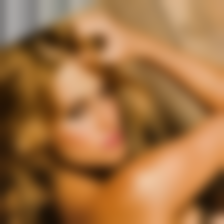}\quad\ic[.17]{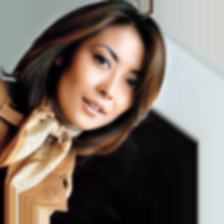}\quad\ic[.17]{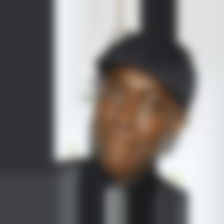}\quad\ic[.17]{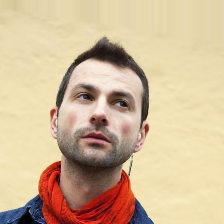}%
  }
  \cap{fig:jittered}{Randomly Perturbed Training Images}{The impact of our random perturbations to image alignment is shown. In \protect\subref*{fig:jittered:no} images are aligned using the corresponding bounding box and angle, which are computed based on hand-labeled facial landmarks, while \protect\subref*{fig:jittered:yes} shows them perturbed with random scale, angle, shift, blur, and horizontal flip.}
\end{figure*}

While making use of the natural alignment of faces, we extend traditional data augmentation techniques to incorporate scaling, rotation, shifting, and blurring of training images.
Given an image annotated with attribute labels and some facial landmarks \cite{liu2015deep}, we align the face after applying random modifications to the scale, the angle, and the location of the bounding box.
Specifically, given the labels of the two eyes $\vec t_{e_r}, \vec t_{e_l}$ and the mouth corners $\vec t_{m_r}, \vec t_{m_l}$ with $\vec t = (x,y)^\mathrm{T}$, we compute the center of both and their respective distance $d$:
\begin{equation}
  \label{eq:landmark-center}
  \vec t_e = \frac{\vec t_{e_r} + \vec t_{e_l}}2, \ \ \ \vec t_m = \frac{\vec t_{m_r} + \vec t_{m_l}}2, \ \ \  d = \|\vec t_e - \vec t_m\|\,,
\end{equation}
then we estimate the bounding box coordinates (specifying left, top, right and bottom coordinates $x_l, y_t, x_r, y_b$):
\begin{equation}
  \label{eq:bbx-coordinates}
  \begin{aligned}
    x_l &= x_e - 0.5\cdot w & x_r &= x_e + 0.5\cdot w \\
    y_t &= y_e - 0.45\cdot h & y_b &= y_e + 0.55\cdot h
  \end{aligned}
\end{equation}
as well as the rotation angle obtained from the eye locations:
\begin{align}
  \label{eq:bbx-angle}
  \alpha = \atan \frac{y_{e_r} - y_{e_l}}{x_{e_r} - x_{e_l}}\,.
\end{align}
The height $h$ and width $w$ of the bounding box are estimated based on the mouth-eye-distance: $h = w = 5.5\cdot d$.

We have implemented a new data layer in Caffe \cite{jia2014caffe} that randomly perturbs the bounding boxes of images in each training epoch \emph{on the fly}, using Bob \cite{anjos2012bob} to handle image transformations and OpenMP \cite{dagum1998openmp} for parallelization on the CPU.
In this layer, an original image and its pre-computed bounding box and angle $\alpha$ are loaded, the scale $s=W/w = H/h$ is estimated, offsets for angle $r_\alpha\sim\mathcal N_{0,20}$, shift $r_y, r_x \sim\mathcal N_{0,0.05}$, and scale $r_s \sim \mathcal N_{1,0.1}$ are randomly drawn and added to the coordinates:
\begin{equation}
  \label{eq:bbx-random}
  \begin{aligned}
    \!\tilde x_l & = x_l + \tilde r_x \cdot w, & \tilde x_r & = x_r + \tilde r_x \cdot w, & \tilde \alpha &= \alpha + r_\alpha, \\
    \!\tilde y_t & = y_t + \tilde r_y \cdot w, & \tilde y_b & = y_b + \tilde r_y \cdot w, & \tilde s &= s \cdot \tilde r_s.
  \end{aligned}
\end{equation}
Using these coordinates, the image is rotated, scaled, and cropped into an RGB image with resolution $W=H=224$, and horizontally flipped with 50\,\% probability.
To emulate smaller image resolutions, yielding blurred upscaled images, a Gaussian filter with a random standard deviation of $\sigma \sim\mathcal N_{0,3}$ is applied to smooth the image, which is finally fed as input to the network training.
\fig{jittered} shows some examples of training images cropped with and without random perturbations.
These random perturbations are much larger than used in related work \cite{rowley1998neural,maronidis2011improving,paulin2014transformation}.
Note that all parameters have been selected based on the characteristics of the CelebA dataset and the facial attribute classification task.
Particularly, the amount of random scale, angle, shift, and blur are selected to work with the rotations of the faces in the CelebA dataset, as well as with the face detector \cite{ranjan2016hyperface}.
However, these parameters can easily be adapted for other image resolutions, contents, and face detectors.

\subsection{Data Augmentation for Testing}
\label{sec:cropping}

Data augmentation can also effectively be used for testing in order to further enhance performances of DCNNs~\cite{howard2014some}.
A common practice is to combine predictions of ten transformations into a final prediction -- i.e., using their average -- by rescaling the image and taking the center crop together with crops of the four corners of the original and the horizontally flipped images.
In our experiments, we adopt this strategy and rescale the test images to a resolution of $256\times256$ pixels before taking those ten crops.
We average the resulting DCNN output scores per attribute, and threshold the results at $\tau=0$ to obtain the final attribute classification.

Since we want to train our network in a way that recognition is stable against rotations, scales, and shifts, we also apply a different augmentation technique, which is unique and has not been reported in the literature.
Instead of taking only crops, yielding only different shifts in the image, we use all combinations of shifts $r_x, r_y \in \{-10, 0, 10\}$, scales $r_s \in \{0.9, 1., 1.1\}$, and angles $r_a \in \{-10, 0, 10\}$, as well as the mirrored versions of these.
In total, using \eqref{eq:bbx-random} we apply 162 different transformations to detected bounding boxes.

%% file: experiments.tex
\section{Experiments}

\subsection{Dataset}

We conduct our experiments on the CelebA dataset \cite{liu2015deep}, which consists of more than 200k images, with an average of 20 images of approximately 10k celebrities.
Most images show faces with a close-to-frontal pose, but some half-profile and few profile faces are also included.
The training set is formed by images of the first 8k identities and contains 160k images.
The remaining 2k identities are equally split into validation and test sets, each having approximately 20k images.
The dataset provides five hand-labeled key-points -- both eyes, the mouth corners, and the nose tip -- and binary labels of 40 facial attributes for each image.

To perform a fair comparison with the LNets+ANet \cite{liu2015deep} approach that did not use hand-labeled annotations, we use the state-of-the-art face detector \cite{ranjan2016hyperface} on the full size images of the CelebA test set.
In some of our experiments, we use the automatically provided landmark locations to align faces, while later we take only the detected bounding boxes, which we enlarge to approximately fit to the content used for training, see Fig.~\subref{fig:jittered:no}, and crop the images to resolution $224\times224$ pixels.
In around 22\,\% of the cases, the overlap of the detected faces with the ground-truth defined in \eqref{eq:bbx-coordinates} is less than 50\,\%.
For the few images with multiple detections, we use the one with the largest overlap with the box derived from the ground-truth key-points.
When the face detector fails, which happens in 40 of the 19962 CelebA test images, we scale the smaller side of the image to 224 pixels and take the center crop -- regardless of the actual size of the face and whether it is the middle of the image.

\subsection{Attribute Classification on Aligned Images}

\input{errors}

We commence our experiments with training and comparing residual networks using images that are aligned using the hand-annotated landmarks, see \sfig{jittered:no} for examples.
We do not incorporate any form of data augmentation during training.
We train ResNet-50 models using Euclidean or sigmoid cross-entropy loss using the same hyper-parameters.
The batch size is set to 64 images and, hence, one epoch requires approximately 2500 iterations.
We choose an RMSProp update rule with inverse learning rate decay policy.
Starting with a learning rate of $\lambda=10^{-3}$, we manually adjust $\lambda$ by a factor of 0.1 two times when the validation set loss stops decreasing.

After training three networks with each loss function, we select the network with the lowest loss on the validation set.
ResNet-50 models trained with either Euclidean (ResNet-50-E/A in \tab{Errors}) or sigmoid cross-entropy loss (not shown in \tab{Errors}) lead to better performing networks than the current state-of-the-art MOON on images aligned (A) using the hand-labeled facial landmarks.
We observe that with 8.49\,\% overall error, ResNet-50-S/A slightly outperforms the 8.64\,\% of ResNet-50-E/A, and for both, the classification error is significantly better than MOON's 9.06\,\%.

It is well known that initializing a network with transferred features -- even from very dissimilar tasks -- can result in better generalization than starting from random network weights \cite{yosinski2014transferable}.
Therefore, we fine-tune ResNets initialized with features of the ResNet-50 model optimized on the ImageNet dataset\footnote{We obtained the pre-trained 50-layer residual network from:\\{\scriptsize\url{http://github.com/KaimingHe/deep-residual-networks}}} by adding an extra fully connected layer with 40 outputs, yielding our ResNet-51.
For comparison, we fine-tune ResNet-51 networks with Euclidean and sigmoid cross-entropy loss.
We use an initial learning rate of $\lambda=10^{-4}$, otherwise we apply the same hyper-parameters as before.
Similarly to previous experiments, we train three networks with each objective function and select the one with the lowest loss on the validation set.
While optimizing with Euclidean loss leads to 8.17\,\% overall error rate (not in \tab{Errors}), applying sigmoid loss results in our best performing single network (ResNet-51-S/A in \tab{Errors}) with 8.16\,\% classification error rate on the CelebA dataset.

To further advance the overall performance, we conduct experiments using ensembles of networks and applying test data augmentation.
We observe that by forming ensembles of our three fine-tuned residual networks -- optimized either with Euclidean or sigmoid cross-entropy loss -- we can improve over the best classification error rate that we achieved with a single DCNN.
The results are very similar for both loss functions: 8.00\,\% with Euclidean (ResNet-51-E Ensemble/A in \tab{Errors}), and 8.01\,\% overall error rate using sigmoid cross-entropy loss (not in \tab{Errors}), both yielding 11.7\,\% relative improvement over MOON.
Considering the gain over single networks, the application of ensembles has a good potential to boost the overall performance -- even though all three networks in the ensemble have the same topology and are trained on the same data.

For test data augmentation, we rescale test images to $256\times256$ pixels, extract five crops (C), horizontally flip them, and use the average network output of the 10 crops as the final prediction.
Surprisingly, we experience a significant degradation in classification performance as error rates increase to \textgreater10\,\%.
For example, ResNet-51-S/C in \tab{Errors} suffers an approximately 2\,\% degradation in overall error rate when tested on ten crops, and ResNet-50-E/C even increases to 10.35\,\% error (not in \tab{Errors}).
Astonishingly, even the smallest data transformations -- using $225\times225$ pixel images for initial rescaling -- raise error rates.

In the previous experiments, test images have been aligned using hand-labeled annotations provided by the dataset \cite{liu2015deep}.
To see how well ResNet-51-S performs on more realistic images, we use the landmarks automatically localized (L) by Ranjan \etal{ranjan2016hyperface} to align images.
The results are labeled as ResNet-51-S/L in \tab{Errors}.
As compared to images aligned using ground-truth landmarks (ResNet-51-S/A), error rates increased for each and every attribute.
Particularly, attributes that require precise alignment such as \textit{Mouth Slightly Open}, \textit{Smiling} or \textit{Wearing Necktie} are impacted.
While this phenomenon can be explained as the result of training with aligned images, it also highlights that these DCNNs are sensitive to small misalignment and their performances highly depend on landmark detection.

\subsection{Face Alignment? Who Cares!}
\label{sec:who-cares}

What we need is a network that is more stable against reasonable variations in scale, angle, and location of the face in the extracted image, and also capable of extracting attributes from a range of image resolutions.
To achieve this goal, we take our best performing network, i.e., ResNet-51-S, but this time we train it using the data augmentation technique described in \sec{jittering}, resulting in the AFFACT network.
Despite the training set augmentation,\footnote{Technically, we do not perform data \emph{augmentation} per se, as we never use the originally aligned images, e.g., from \sfig{jittered:no} during training, but only perturbed images as shown in \sfig{jittered:yes}.} we apply the exact same training technique as before.
For the validation set, instead of using aligned faces, we employ images cropped based on bounding boxes, which are computed from the hand-labeled annotations using \eqref{eq:bbx-coordinates} without random perturbations and $\alpha=0$.

In order to verify that we do not give up on accuracy on the aligned faces, we test our AFFACT network on the aligned test images.
The results are listed as AFFACT/A in \tab{Errors}.
As we can see, compared to the base ResNet-51-S, the overall error rate slightly increases from 8.16\,\% to 8.33\,\%.
For most attributes the error increases with the same amount, but, interestingly, for some attributes that are affected by larger parts of the face such as \textit{Blond Hair}, \textit{Chubby}, \textit{Male}, or \textit{Young}, AFFACT/A can even achieve better results than ResNet-51-S/A.

The goal of training with our data augmentation technique is to make the AFFACT network less dependent on face alignment.
Ideally, this would help attribute classification in two different ways, i.e., we would be able to use image crops that have failed miserably for ResNet-51, and we could use faces detected by a face detector.
Indeed, when applying the same ten crops, the AFFACT network improves accuracy -- the error rates given as AFFACT/C in \tab{Errors} are comparable to the ones of ResNet-51-S on aligned images without taking crops.
The most impressive characteristic of the AFFACT network, however, is its robustness to facial misalignment.
We test how well AFFACT can handle images that are aligned using automatically detected landmarks \cite{ranjan2016hyperface}.
The results presented as AFFACT/L in \tab{Errors} show that AFFACT can handle localization errors much better than ResNet-51-S.
While for the latter the error rate increases from 8.16\,\% to 9.07\,\% when hand-labeled or automatically detected landmarks are used, AFFACT results remain more stable with error rates only increasing from 8.33\,\% to 8.47\,\%.

Now, we compare ResNet-51-S and AFAFCT using only the face detections (D) obtained with the state-of-the-art face detector from Ranjan \etal{ranjan2016hyperface}.
These results are now directly comparable with the LNets+ANet approach \cite{liu2015deep}, which did not use facial landmarks and reached an average classification error of 12.70\,\% \cite{rudd2016moon}.
As seen in \tab{Errors}, the results of both networks deteriorate, which is mainly due to the fact that some of the faces were not detected correctly, i.e., parts of the faces were not inside the bounding boxes.
Although the 9.57\,\% error rate of ResNet-51-S/D is already better than the 12.70\,\% of LNets+ANet, this is radically outperformed by 8.55\,\% of AFFACT/D.
Looking at the errors of various attributes, we can observe that on detected faces the main advantage of AFFACT over ResNet-51-S is in highly localized attributes, such as \textit{Arched Eyebrows}, \textit{Bushy Eyebrows}, \textit{Heavy Makeup}, \textit{Mouth Slightly Open}, or \textit{Smiling}.
Furthermore, there is not a single attribute that AFFACT classifies worse than LNets+ANet.
Interestingly, our AFFACT results using only the detected bounding boxes also outperform the old state-of-the-art MOON network, which required images to be aligned using hand-labeled annotations.

As we have seen before, AFFACT can handle multiple crops of an image.
Hence, we apply the same ten crops (C) on images with detected bounding boxes (D), we obtain a slight reduction of error, i.e., from 8.55\,\% without crops to 8.45\,\% with crops (AFFACT/CD in \tab{Errors}).
AFFACT is designed to work with different scales and rotation angles, so there is no need to limit ourselves to the traditional ten crops, thus we can use many more transformations (T) including different rotation angles, scales, shifts, and mirroring of the image as described in \sec{cropping}.
When applying all 162 transformations and thresholding the averaged network predictions, we can gain an enormous boost in average classification performance, i.e., the error drops further to 8.09\,\% (AFFACT/TD in \tab{Errors}).
Interestingly, most of the boost in performance is obtained by a few attributes like \textit{Big Lips} or \textit{Narrow Eyes}, which have been predicted quite poorly before, while for some attributes such as \textit{Brown Hair} or \textit{Goatee} the error increased considerably.

Finally, when combining all methods investigated in this paper, i.e., using an ensemble of three AFFACT networks and averaging over 162 transformations of the detected bounding boxes, we can lower the classification error to 8.03\,\% (AFFACT-Ensemble/TD in \tab{Errors}), which represents the new state of the art on the CelebA benchmark using detected faces without alignment with a relative improvement of 36.8\,\% over LNets+ANet~\cite{liu2015deep}.
Interestingly, the boost in performance over using 162 transformations of one network is minor and statistically not very significant ($p \approx 0.03$ in a two-sided paired t-test over a large sample size).
Hence, using the transformations of images in one or more networks does not make a difference, so we can save the training and extraction time, as well as the memory for the network ensemble, and use only a single AFFACT network.

%% file: errors.tex
\newcommand*\rot{\rotatebox{90}}
\newcommand\cen[1]{\multicolumn{1}{c}{#1}}
\newcommand\cenl[1]{\multicolumn{1}{c|}{#1}}
\newcommand\pb[1]{\parbox{1.29cm}{\centering #1}}

\definecolor{darkgreen}{RGB}{0,64,64}
\renewcommand\r{\color{red}\bf}
\newcommand\g{\color{green}\bf}
\renewcommand\b{\color{blue}\bf}
\newcommand\m{\color{magenta}\bf}

\begin{table*}[t!]
  \renewcommand\arraystretch{1.03}
  \scriptsize
  \centering
  \begin{tabular}{r||r|rrrr|rr|rr|rrrrr|r}
    \toprule
   	& \cenl{\rot{\pb{MOON}}}
   	& \cen{\rot{\pb{ResNet-50-E}}}
   	& \cen{\rot{\pb{ResNet-51-S}}}
   	& \cen{\rot{\pb{ResNet-51-E \linebreak Ensemble}}}
   	& \cenl{\rot{\pb{AFFACT}}}
   	& \cen{\rot{\pb{ResNet-51-S}}}
   	& \cenl{\rot{\pb{AFFACT}}}
   	& \cen{\rot{\pb{ResNet-51-S}}}
   	& \cenl{\rot{\pb{AFFACT}}}
   	& \cen{\rot{\pb{ResNet-51-S}}}
   	& \cen{\rot{\pb{AFFACT}}}
   	& \cen{\rot{\pb{AFFACT}}}
   	& \cen{\rot{\pb{AFFACT}}}
   	& \cenl{\rot{\pb{AFFACT \linebreak Ensemble}}}
   	& \cen{\rot{\pb{LNets+ANet}}}\\

    \midrule
    Tested on & \cen{A} & \cen{A} & \cen{A} & \cen{A} &\cenl{A} & \cen{C} & \cenl{C} & \cen{L} & \cenl{L} & \cen{D} & \cen{D} & \cen{CD} & \cen{TD} & \cenl{TD} & \cen{D*}\\

    \midrule\midrule

    \input{errordata}

    \bottomrule

  \end{tabular}
  \cap{tab:Errors}{Attribute Classification Errors}{The error rates for each attribute are shown separately. Networks are tested on: (A) images aligned with hand-labeled annotations, (C) 10 crops of aligned images, (L) images aligned with automatically detected landmarks, (D) detected face bounding boxes, and (T) 162 transformations of detected bounding boxes. For comparison, the old state-of-the-art results of MOON \cite{rudd2016moon} and LNets+ANet \cite{liu2015deep} are given on either side of the table. The best values per block are highlighted in color.}
\end{table*}

%% file: errordata.tex
5 o Clock Shadow & 5.97 & 5.64 & \r 4.94 & 5.07 & 5.13 & 7.90 & \b 4.97 & 6.27 & \m 5.25 & 5.72 & \g 5.19 & 5.23 & 5.70 & 5.62 & 9 ~\\
Arched Eyebrows & 17.74 & 16.16 & 14.65 & \r 14.54 & 15.79 & 22.01 & \b 15.69 & 16.36 & \m 16.06 & 20.19 & 16.70 & 15.89 & 14.48 & \g 14.45 & 21 ~\\
Attractive & 18.33 & 17.26 & 16.63 & \r 16.39 & 17.14 & 19.48 & \b 16.97 & 17.37 & \m 17.15 & 20.25 & \g 17.20 & 17.30 & 18.38 & 18.63 & 19 ~\\
Bags Under Eyes & 15.08 & 14.72 & 14.08 & \r 13.95 & 14.69 & 19.10 & \b 14.34 & 15.06 & \m 14.83 & 17.30 & 15.09 & \g 14.87 & 15.78 & 15.77 & 21 ~\\
Bald & 1.23 & 1.16 & 0.94 & \r 0.89 & 0.91 & 1.06 & \b 0.97 & 1.12 & \m 0.93 & 1.04 & \g 0.96 & 1.00 & 1.06 & 0.98 & 2 ~\\
Bangs & 4.20 & 4.14 & 3.77 & \r 3.74 & 3.90 & 4.42 & \b 3.88 & \m 3.90 & 3.93 & 4.47 & 3.99 & \g 3.88 & 4.54 & 4.49 & 5 ~\\
Big Lips & 28.52 & 27.99 & 27.53 & \r 27.27 & 27.30 & 30.04 & \b 27.54 & 28.49 & \m 27.48 & 28.81 & 28.32 & 28.25 & 16.74 & \g 16.00 & 32 ~\\
Big Nose & 16.00 & 15.65 & 14.82 & \r 14.51 & 15.62 & 16.46 & \b 15.26 & 16.37 & \m 15.57 & 16.27 & 15.43 & \g 15.18 & 16.99 & 16.98 & 22 ~\\
Black Hair & 10.60 & 10.33 & \r 9.59 & 9.71 & 9.64 & 10.29 & \b 9.45 & 10.24 & \m 9.53 & 10.49 & 9.73 & 9.64 & \g 8.39 & 8.43 & 12 ~\\
Blond Hair & 4.14 & 4.02 & 4.09 & 4.00 & \r 3.80 & 3.91 & \b 3.81 & 4.08 & \m 3.85 & 4.15 & 3.96 & \g 3.76 & 4.26 & 4.35 & 5 ~\\
Blurry & 4.33 & 3.88 & 3.64 & \r 3.49 & 3.63 & 6.97 & \b 3.60 & 5.51 & \m 3.97 & 4.81 & \g 3.90 & 4.01 & 4.03 & 3.91 & 16 ~\\
Brown Hair & 10.62 & 10.29 & 10.61 & \r 10.23 & 11.60 & \b 10.62 & 10.88 & \m 10.61 & 11.50 & \g 11.24 & 11.92 & 11.47 & 14.55 & 14.26 & 20 ~\\
Bushy Eyebrows & 7.38 & 7.06 & 7.27 & \r 7.02 & 7.54 & 11.33 & \b 7.55 & 7.72 & \m 7.69 & 10.57 & 7.52 & 7.33 & 7.65 & \g 7.21 & 10 ~\\
Chubby & 4.56 & 4.47 & 4.37 & \r 4.09 & 4.26 & 4.47 & \b 4.17 & 6.08 & \m 4.34 & 4.73 & 4.30 & \g 4.23 & 4.40 & 4.27 & 9 ~\\
Double Chin & 3.68 & 3.65 & 3.58 & \r 3.37 & 3.48 & 3.66 & \b 3.57 & 4.79 & \m 3.58 & 3.94 & 3.57 & 3.52 & 3.24 & \g 3.23 & 8 ~\\
Eyeglasses & 0.53 & 0.34 & 0.32 & \r 0.29 & 0.39 & 0.66 & \b 0.32 & \m 0.35 & 0.40 & 0.55 & 0.41 & \g 0.39 & 0.56 & 0.51 & 1 ~\\
Goatee & 2.96 & 2.57 & 2.29 & \r 2.24 & 2.50 & 2.91 & \b 2.45 & 2.72 & \m 2.51 & 2.66 & 2.44 & \g 2.43 & 3.29 & 3.26 & 5 ~\\
Gray Hair & 1.90 & 1.71 & \r 1.58 & 1.62 & 1.69 & \b 1.59 & 1.61 & 1.78 & \m 1.66 & 1.73 & 1.70 & \g 1.65 & 1.99 & 1.90 & 3 ~\\
Heavy Makeup & 9.01 & 8.45 & 7.83 & \r 7.77 & 7.90 & 9.53 & \b 7.73 & 8.66 & \m 7.96 & 11.06 & 8.17 & \g 8.13 & 8.15 & 8.33 & 10 ~\\
High Cheekbones & 12.99 & 12.37 & 11.90 & \r 11.73 & 12.23 & 21.52 & \b 12.27 & 13.35 & \m 12.37 & 15.32 & 12.77 & 12.44 & \g 11.55 & 11.65 & 13 ~\\
Male & 1.90 & 1.93 & 1.66 & \r 1.50 & \r 1.50 & 2.58 & \b 1.49 & 2.05 & \m 1.78 & 2.66 & 1.77 & 1.73 & \g 1.31 & 1.34 & 2 ~\\
Mouth Slightly Open & 6.46 & 6.21 & \r 5.76 & 5.93 & 5.94 & 12.13 & \b 5.78 & 9.25 & \m 6.18 & 9.09 & 6.22 & \g 6.04 & 6.17 & 6.11 & 8 ~\\
Mustache & 3.18 & 2.95 & \r 2.73 & 2.79 & 2.89 & 3.63 & \b 2.89 & 3.39 & \m 3.04 & 3.22 & \g 2.86 & 2.98 & 3.50 & 3.56 & 5 ~\\
Narrow Eyes & 13.48 & 12.42 & \r 12.12 & 12.22 & 12.31 & 12.86 & \b 12.29 & 12.63 & \m 12.42 & 13.47 & 12.60 & 12.51 & 6.30 & \g 6.23 & 19 ~\\
No Beard & 4.42 & 3.93 & 3.53 & \r 3.36 & 3.55 & 5.50 & \b 3.46 & 4.24 & \m 3.84 & 4.31 & \g 3.60 & 3.65 & 4.04 & 4.04 & 5 ~\\
Oval Face & 24.27 & 23.19 & 22.12 & \r 21.66 & 22.59 & 26.00 & \b 22.69 & 24.71 & \m 23.26 & 25.92 & 25.22 & 24.35 & \g 23.09 & 23.25 & 34 ~\\
Pale Skin & 3.00 & 2.91 & \r 2.75 & 2.85 & 2.95 & 3.00 & \b 2.80 & \m 2.92 & 2.94 & 2.93 & 2.82 & \g 2.81 & 3.25 & 3.25 & 9 ~\\
Pointy Nose & 23.54 & 22.69 & 22.43 & \r 21.96 & 23.10 & 24.84 & \b 22.62 & \m 22.69 & 22.88 & 24.18 & 22.88 & 22.65 & 22.60 & \g 22.39 & 28 ~\\
Receding Hairline & 6.44 & 6.29 & \r 5.92 & 5.95 & 6.33 & 6.35 & \b 6.17 & 7.41 & \m 6.35 & 7.07 & 6.32 & 6.14 & 5.24 & \g 5.10 & 11 ~\\
Rosy Cheeks & 5.18 & 4.83 & 4.60 & \r 4.46 & 4.84 & 6.95 & \b 4.82 & \m 4.80 & \m 4.80 & 6.03 & 5.04 & \g 4.82 & 4.87 & \g 4.82 & 10 ~\\
Sideburns & 2.41 & 2.26 & 1.93 & \r 1.91 & 2.16 & 2.20 & \b 2.09 & 2.22 & \m 2.17 & 2.10 & 2.14 & \g 2.05 & 2.67 & 2.69 & 4 ~\\
Smiling & 7.40 & 7.12 & 6.83 & \r 6.51 & 7.04 & 13.54 & \b 6.65 & 9.91 & \m 7.22 & 11.79 & 7.22 & 7.15 & 7.20 & \g 7.11 & 8 ~\\
Straight Hair & 17.74 & 16.32 & 14.80 & \r 14.19 & 14.74 & 15.76 & \b 14.69 & 15.62 & \m 15.00 & 15.03 & 14.77 & 14.76 & \g 14.35 & 14.45 & 27 ~\\
Wavy Hair & 17.53 & 17.05 & 14.70 & \r 13.69 & 13.75 & 15.81 & \b 14.04 & 15.66 & \m 14.35 & 15.07 & 13.93 & 14.13 & 12.41 & \g 12.12 & 20 ~\\
Wearing Earrings & 10.40 & 9.74 & 9.02 & \r 8.84 & 9.02 & 10.53 & \b 8.95 & 9.24 & \m 9.03 & 9.55 & 9.08 & 9.05 & 8.06 & \g 7.97 & 18 ~\\
Wearing Hat & 1.05 & 0.98 & \r 0.82 & 0.84 & 0.90 & 0.85 & \b 0.82 & 1.00 & \m 0.88 & 1.02 & \g 0.86 & 0.88 & 1.02 & 1.06 & 1 ~\\
Wearing Lipstick & 6.07 & 6.03 & 5.93 & \r 5.62 & 6.04 & 8.41 & \b 6.06 & 6.54 & \m 6.28 & 8.16 & \g 5.95 & 6.03 & 7.34 & 7.26 & 7 ~\\
Wearing Necklace & 12.96 & 11.85 & 10.59 & \r 10.40 & 10.73 & 12.09 & \b 11.39 & \m 11.57 & 11.67 & 10.86 & 11.16 & 11.68 & 9.77 & \g 9.71 & 29 ~\\
Wearing Necktie & 3.37 & 3.04 & \r 2.71 & 2.72 & \r 2.71 & 3.92 & \b 2.77 & 4.21 & \m 3.06 & 2.85 & \g 2.84 & 2.87 & 3.26 & 3.23 & 7 ~\\
Young & 11.92 & 11.89 & 11.17 & \r 10.83 & 11.02 & 11.88 & \b 11.01 & 11.88 & \m 11.15 & 12.12 & 11.35 & \g 11.27 & 11.42 & 11.28 & 13 ~\\
\midrule\midrule
OVERALL & 9.06 & 8.64 & 8.16 & \r 8.00 & 8.33 & 10.17 & \b 8.26 & 9.07 & \m 8.47 & 9.57 & 8.55 & 8.45 & 8.09 & \g 8.03 & 12.70 ~\\

%% file: conclusion.tex
\section{Conclusions}

\begin{figure}
  \centering
  \subfloat[\label{fig:validation:ResNet-50}ResNet-50-S]{\includegraphics[width=.43\textwidth]{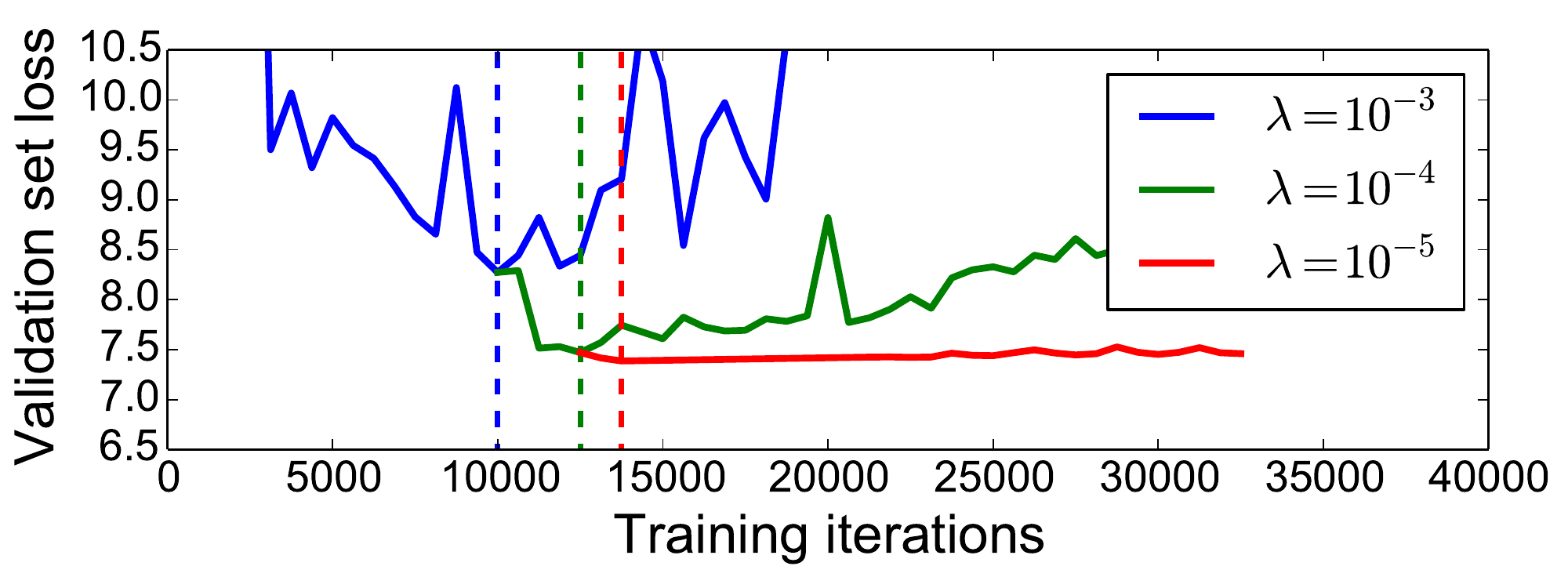}}

  \subfloat[\label{fig:validation:ResNet-51}ResNet-51-S]{\includegraphics[width=.43\textwidth]{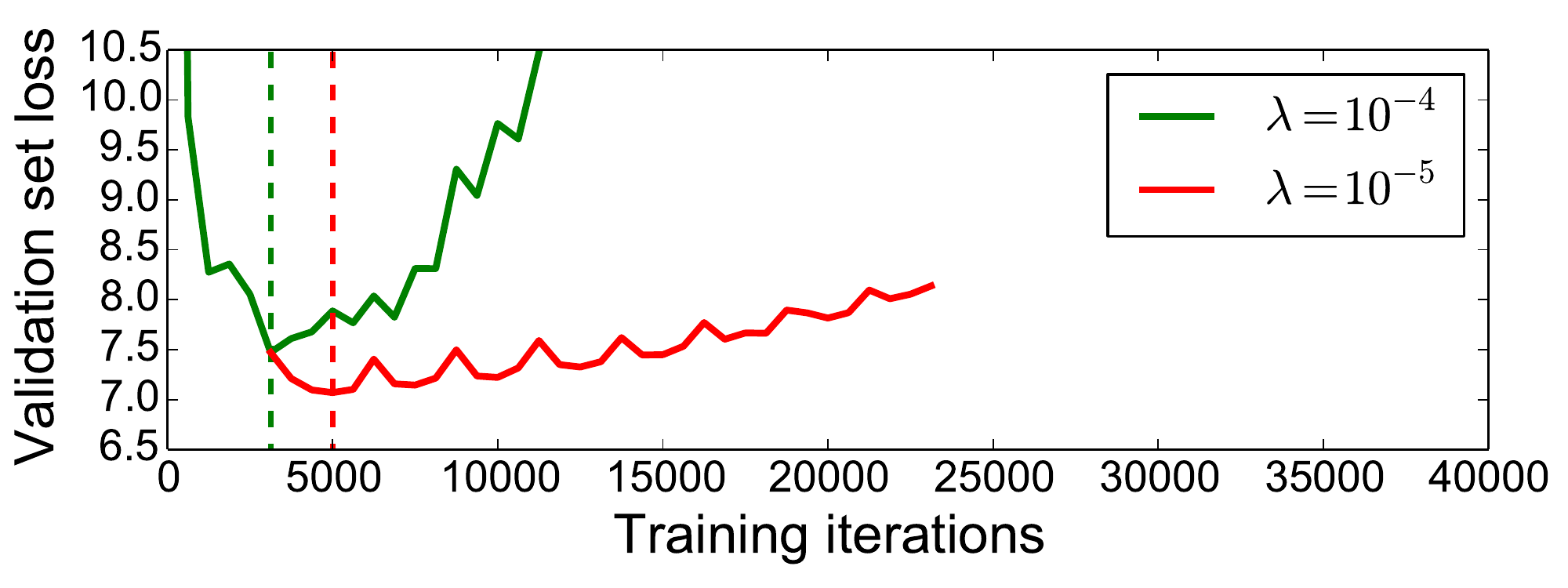}}

  \subfloat[\label{fig:validation:AFFACT}AFFACT]{\includegraphics[width=.43\textwidth]{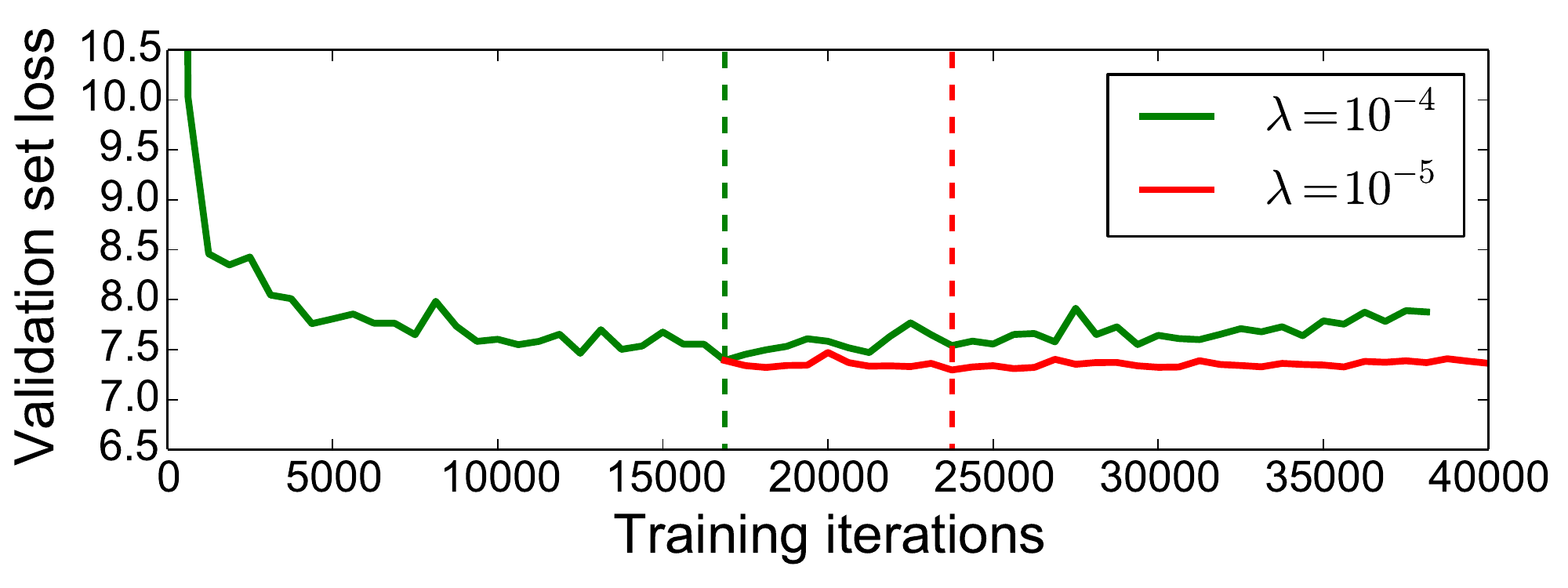}}

  \cap{fig:validation}{Validation set Losses}{This figure shows the trends of the validation loss obtained during the multi-stage training of \protect\subref*{fig:validation:ResNet-50} ResNet-50-S, \protect\subref*{fig:validation:ResNet-51} ResNet-51-S, and \protect\subref*{fig:validation:AFFACT} AFFACT. Dotted lines mark the lowest validation set loss for each learning rate.}
\end{figure}

In our experiments, we obtain the new state-of-the-art facial attribute classification on the CelebA \cite{liu2015deep} benchmark with classification errors of 8.00\,\% on aligned images and 8.03\,\% on detected faces.
The achieved relative improvements over the previous state of the art are 11.7\,\% over MOON \cite{rudd2016moon} and 36.8\,\% over LNets+ANet \cite{liu2015deep}, respectively.
The results on aligned faces are achieved by using an ensemble of three ResNet-51-S networks that are pre-trained on the ImageNet image classification task and fine-tuned on the CelebA training set.
When testing two different loss functions -- sigmoid cross-entropy loss and Euclidean loss -- we have found little difference in classification errors.

To obtain the results on detected faces, we implemented a data augmentation technique and trained the AFFACT network.
We showed that AFFACT can use the bounding boxes provided by a face detector without further alignment to classify facial attributes.
To our best knowledge, we present the first experiments showing that a face processing task can be done using only the detected bounding boxes and still get the same performance as with facial landmark localization.
One might argue that the attribute prediction of AFFACT on automatically detected landmarks (AFFACT/L in \tab{Errors}) is still better than using bounding boxes (AFFACT/D).
However, when performing a two-sided paired t-test using attribute classification errors of AFFACT/L and AFFACT/D, the p-value of \textgreater{}0.23 does not indicate a statistically significant difference.
Hence, we can conclude that, statistically, there is no improvement by using automatically detected landmarks over using the detected bounding boxes.
Also, we recognize that the CelebA dataset provides images in rather controlled environments.
When taking into account more difficult data such as provided in the IJB-A dataset \cite{klare2015ijba}, chances of mis-detected facial landmarks increase and, therewith, more images will be misaligned.
Last but not least, our data augmentation technique is generic and can be easily adapted to other face processing tasks, e.g., face recognition.
We will investigate this in our future work.

As we artificially increase the training set size by our data augmentation technique and make images harder to be classified, training time naturally increases.
In \fig{validation}, we show the validation set errors that we obtained during our multi-stage training of ResNet-50-S, ResNet-51-S and AFFACT.
When training ResNet-50-S from scratch on aligned images, it took around 14k iterations with a batch-size of 64.
When fine-tuning ResNet-51-S from the pre-trained ImageNet model, only 5k iterations were required until convergence on the validation set, which is around 2 training epochs.
Adding data augmentation, the number of iterations again increased to 24k iterations, i.e., close to 10 epochs.
Interestingly, this is still lower than the 24 epochs that Rudd \etal{rudd2016moon} reported for training their MOON network.

In \sec{who-cares}, we experimented with computing average predictions for multiple crops of images in order to improve facial attribute classification.
While for ResNet-51-S, which was trained on aligned images, the classification error increased, the AFFACT network was able to utilize several crops of the face to further improve attribute classification.
The best performance was obtained with 162 transformations of different rotations, scales, shifts, and mirrored test images.
However, the attribute classification only improved for some attributes while for others the results deteriorated.
Therefore, running a two-sided paired t-test between AFFACT/CD and AFFACT Ensemble/TD from \tab{Errors} results in a p-value of \textgreater{}0.39, which does not support a conclusion of a significant difference.
Interestingly, using ensembles of networks improves accuracy significantly when the networks are presented with a single image without taking crops.
However, when 10 crops or 162 perturbations are used, the improvement of the ensemble network is small and not statistically significant.
In this paper, we have fused the 162 predictions only with averaging;  other fusion techniques such as majority voting or outlier detection might further improve accuracies.
Also, each perturbation requires a full network evaluation and reducing the number of perturbations will increase usability.
We leave these for future research.

As indicated by Rudd \etal{rudd2016moon}, training end-to-end DCNNs using a biased dataset influences the distributions of positive and negative classification errors, which surely also affects the AFFACT network.
How well our AFFACT attribute predictions work in a face recognition task, and whether the proposed training and test augmentation techniques would be useful for that task will be addressed by future work.
Preliminary experiments show that AFFACT attributes can outperform MOON's for face recognition on LFW, but only when trained with Euclidean loss.

Lately, we have realized that a lot of images in the CelebA dataset have incorrect attribute labels.
We are currently relabeling some of those errors.
We assume that real attribute classification errors are at least 25\,\% lower than reported.